\documentclass[letterpaper, 10 pt, conference]{ieeeconf}  

\usepackage{graphicx}
\usepackage{amsmath}
\usepackage{amssymb}   
\usepackage{subcaption}
\usepackage{makecell}
\usepackage{multirow}
\usepackage{xcolor} 
\usepackage{cite}

\newtheorem{lemma}{Lemma}

\usepackage{tabularx}
\usepackage{array}
\newcolumntype{Y}{>{\centering\arraybackslash}X}

\IEEEoverridecommandlockouts                              

\newcommand\defeq{\mathrel{\stackrel{\makebox[0pt]{\mbox{\normalfont\scriptsize def}}}{:=}}}

\title{\LARGE \bf
Cumulative Consensus Score: Label-Free and Model-Agnostic Evaluation of Object Detectors in Deployment
}

\author{Avinaash Manoharan$^{1,2}$, Xiangyu Yin$^{3}$, Domenik Helms$^{1}$, and Chih-Hong Cheng$^{2,3}$%
\thanks{$^{1}$ DLR Institute of Systems Engineering for Future Mobility, Germany}
\thanks{$^{2}$ Carl von Ossietzky University of Oldenburg, Germany}
\thanks{$^{3}$ Chalmers University of Technology,  Sweden}
\thanks{Contact: avinaash.manoharan@dlr.de, chih-hong.cheng@uol.de}
}

\begin{document}

\maketitle
\thispagestyle{empty}
\pagestyle{empty}

\begin{abstract}

Evaluating object detection models in deployment is challenging because ground-truth annotations are rarely available. We introduce the Cumulative Consensus Score (CCS), a label-free monitoring signal for continuous evaluation and comparison of detectors in real-world settings. CCS applies test-time data augmentation to each image and measures the spatial consistency of predicted bounding boxes across augmented views using Intersection over Union. The resulting consensus score serves as a proxy for reliability without requiring bounding box annotations. In controlled experiments on Open Images and KITTI, CCS achieved over 90\% congruence with F1-score, Probabilistic Detection Quality, and Optimal Correction Cost, with 
qualitative consistency further confirmed on COCO and BDD100K across model pairs. The method is model-agnostic, working across single-stage and two-stage detectors, and operates at the case level to highlight under-performing scenarios. We also provide a simplified theoretical link between expected CCS and detection correctness. Altogether, CCS provides a robust foundation for DevOps-style monitoring of object detectors.

\end{abstract}

\section{Introduction}
\label{sec:intro}

Deep learning has enabled major advances in computer vision, particularly in safety-critical domains such as autonomous driving. However, the reliability of perception modules, including object detection, remains a concern. Object detectors can make errors under distribution shifts and inherently suffer from epistemic uncertainty, stemming from incomplete training data and limited coverage of real-world conditions \cite{ABDAR2021243,hall2020probabilistic,Le2018}. This uncertainty makes it difficult for engineers and end users to judge whether a newly trained object detector is more trustworthy than an existing, well-established one. Ground-truth annotations are typically required for supervised metrics (e.g., mAP, F1-score), yet such labels are rarely available at deployment time. This creates a gap between controlled lab evaluation and the operational domain, where continuous monitoring and safe upgrades are most needed.

To address this challenge, we introduce the \emph{Cumulative Consensus Score (CCS)}, a label-free method for evaluating object detectors in deployment. CCS builds on the idea of functional monitoring by enabling direct comparison of a new detector against an existing baseline without requiring annotated data. Its central principle is to assess the \emph{spatial consistency of predictions}: detectors that generalize better are expected to exhibit more stable outputs when the input is subjected to benign transformations. By turning this stability into a measurable signal, CCS provides a practical proxy for reliability. In addition, we provide a simplified theoretical analysis that links CCS to detection correctness under an idealized setting, offering intuition for why spatial consensus reflects reliability.

A key difficulty is that many uncertainty estimation techniques require architectural changes or large ensembles, making comparisons costly. CCS instead uses Test-Time Data Augmentation (TTDA) and requires no additional training. In practice, several photometric variations of an image are processed to produce bounding boxes, and their overlap is quantified using Intersection over Union (IoU). CCS aggregates these overlaps into a single image-level consensus score, enabling comparison between two detectors.

We validate CCS in controlled experiments by comparing its outputs against ground-truth-based metrics including F1 score, Probabilistic Detection Quality (pPDQ), and Optimal Correction Cost (OC-cost). Experiments across datasets such as Open Images~\cite{kuznetsova2020open} and KITTI~\cite{geiger2012ready}, and across architectures including Faster R-CNN~\cite{ren2016fasterrcnnrealtimeobject}, RetinaNet~\cite{lin2017focal} and SSD~\cite{liu2016ssd} show that CCS achieves over $90\%$ congruence with these established measures, with qualitative consistency further confirmed on COCO~\cite{lin2014microsoft} and BDD100K~\cite{yu2020bdd100k}. Importantly, CCS provides results at the image level, enabling the identification of under-performing cases where predictions become unstable across benign transformations allowing engineers to pinpoint problematic inputs and guide targeted improvements.

\section{Related Work}
\label{sec:literature_review}
Several label-free indicators have been explored. Schmidt et al.~\cite{Schmidt2020AdvancedAL} introduced a consensus score in an ensemble-based active learning setting, relying on Bayesian variants~\cite{Kendall2017,wang2019aleatoric} that increase deployment cost. Yang et al.~\cite{Yang2024BoS} proposed the Box Stability Score (BoS) and Yu et al.~\cite{Yu2024DAS} proposed the Detection Adaptation Score (DAS); both correlate stability-related signals with supervised metrics but require internal feature access and mainly target training-time model selection. Yu et al.~\cite{yu2022cald} extended consistency ideas to active learning (CALD), reducing labeling but still relying on labeled retraining. In contrast, CCS operates directly on detector outputs under TTDA, requires neither ensembles nor feature access, and provides a per-image monitoring signal based on augmentation-induced spatial consistency. We further provide a simplified theoretical link between CCS and detection correctness under an idealized setting.

Test-time data augmentation is commonly used to improve robustness and accuracy \cite{Shanmugam2021TTA,Kim2020TTA}; CCS instead uses TTDA to quantify prediction stability as a monitoring signal. Probabilistic object detection explicitly models localization and classification uncertainty~\cite{gasperini2021certainnet,Feng2022,nallapareddy2023evcenternet}, but typically requires architectural modifications or ensembles and is detector-specific. Ground-truth-based metrics such as pPDQ~\cite{hall2020probabilistic} and OC-cost~\cite{Otani_2022_CVPR} are not applicable without labels. CCS complements these directions by providing a lightweight, model-agnostic proxy for spatial uncertainty that aligns with supervised metrics when labels exist, while remaining directly usable in label-free deployment settings.

\section{Inside Cumulative Consensus Score}

\begin{figure}[t]
  \centering
\includegraphics[width=0.7\columnwidth]{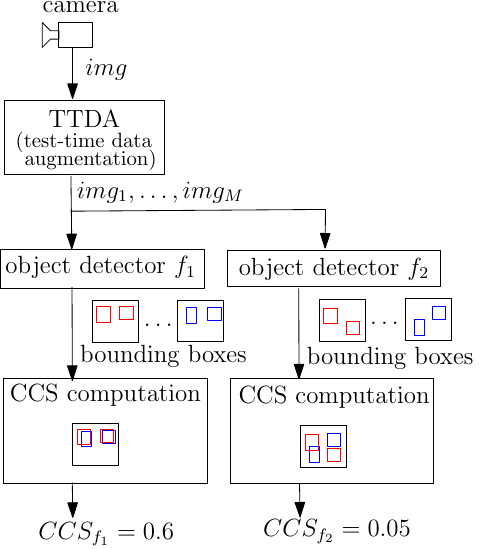}
  \caption{Workflow of comparing the result of two object detectors $f_1$, $f_2$ using the corresponding CCS values.} 
  \label{fig:ccs.computation.workflow}
  \vspace{-5mm}
\end{figure}

This section describes the methodology for computing the Cumulative Consensus Score (CCS) at the image level. Fig.~\ref{fig:ccs.computation.workflow} illustrates the workflow and the mindset for how CCS is designed, as well as comparing the result of two object detectors~$f_1$ and~$f_2$. Given an image $img$, TTDA is conducted to build~$M$ variations using TTDA techniques i.e. photometric augmentations such as shifting weather conditions. As there is no shearing or cropping of images in our augmentation pipeline,  for all augmented images, any object (e.g., Car) should be located at the same location in the image plane. 
In this regard, a \emph{necessary condition} of a high-performing object detector is to generate bounding boxes among images that induce a larger overlap. In Fig.~\ref{fig:ccs.computation.workflow}, object detector~$f_1$ performs better, as the bounding boxes being produced across different images have higher overlap. As shown in subsequent paragraphs, CCS utilizes such a concept and characterizes the degree of overlap by computing an aggregated value utilizing the standard IoU computation across augmented images. 

\begin{figure}[t]
  \centering
\includegraphics[width=\columnwidth]{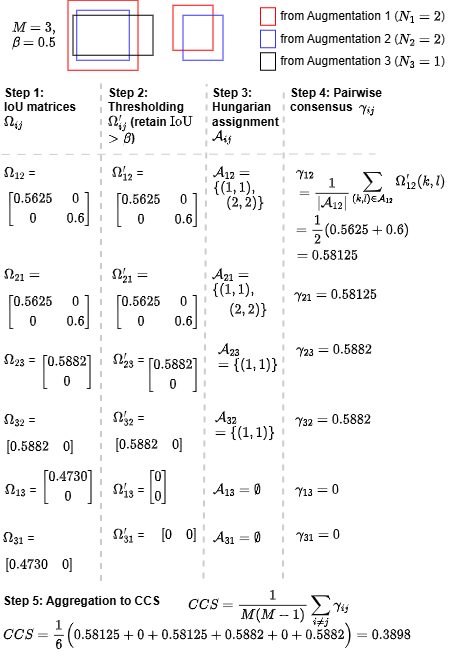}
  \caption{Predictions from three augmentations shown in different coloured boxes and the associated CCS computation.}
  \label{fig:ccs.computation.example}
  \vspace{-5mm}
\end{figure}

\vspace{1mm}

\subsection{Defining CCS}
\subsubsection{Basic setting}

Let $M$ denote the number of augmentations applied to each input image. For a given image, let $N_i$ and $N_j$ denote the number of bounding boxes predicted by the detector on the $i^{\text{th}}$ and $j^{\text{th}}$ augmentations, respectively. 

For each ordered pair $(i,j)$ with $i \neq j$, we compute the Intersection over Union (IoU) matrix
\[
\Omega_{ij} \in \mathbb{R}^{N_i \times N_j},
\]
where each entry $\Omega_{ij}(k,l)$ represents the IoU between the $k^{\text{th}}$ predicted bounding box from augmentation~$i$ and the $l^{\text{th}}$ predicted bounding box from augmentation~$j$. Since $N_i$ and $N_j$ may differ, $\Omega_{ij}$ is not necessarily a square matrix.

As an illustrative example, consider the three augmentations shown in Fig.~\ref{fig:ccs.computation.example}, where the red, blue, and black bounding boxes correspond to augmentations $1$, $2$, and $3$, respectively. For the pair $(1,2)$, the IoU matrix is

\[
\Omega_{12} =
\begin{bmatrix}
0.5625 & 0 \\
0 & 0.6
\end{bmatrix},
\]

indicating that the left red box overlaps strongly with the left blue box (IoU $0.5625$), and the right red box overlaps with the right blue box (IoU $0.6$), while cross-overlaps are zero. Similarly, for the pair $(2,3)$,
\[
\Omega_{23} =
\begin{bmatrix}
0.5882 \\
0
\end{bmatrix},
\]
since augmentation $2$ produces two detections while augmentation $3$ produces only one.

\vspace{1mm}
\subsubsection{Apply thresholding to the IoU matrix}

To suppress weak overlaps, we apply a threshold $\beta$ to obtain a filtered IoU matrix $\Omega'_{ij}$:

\begin{equation}
  \Omega'_{ij}(k,l)  \defeq 
  \begin{cases}
    \Omega_{ij}(k,l), & \text{if } \Omega_{ij}(k,l) > \beta, \\
    0, & \text{otherwise}.
  \end{cases}
  \label{eq:thresholded_iou_matrix}
\end{equation}

Only IoU values strictly greater than $\beta$ are retained. This step removes minor overlaps and emphasizes meaningful spatial agreement. In our example with $\beta = 0.5$, the entries in $\Omega_{12}$ remain unchanged since both non-zero values exceed the threshold, whereas in $\Omega_{13}$ the value $0.4730$ is suppressed to zero, resulting in $\Omega'_{13}$ being identically zero. The choice of~$\beta$ follows the convention established in~\cite{Miller2018icra}, where an IoU of at least~$0.5$ is considered the minimum requirement for associating a detection with a ground-truth object.

\begin{figure*}[!t]
	\centering
	\begin{subfigure}{0.32\textwidth}
		\centering
		\includegraphics[width=\textwidth]{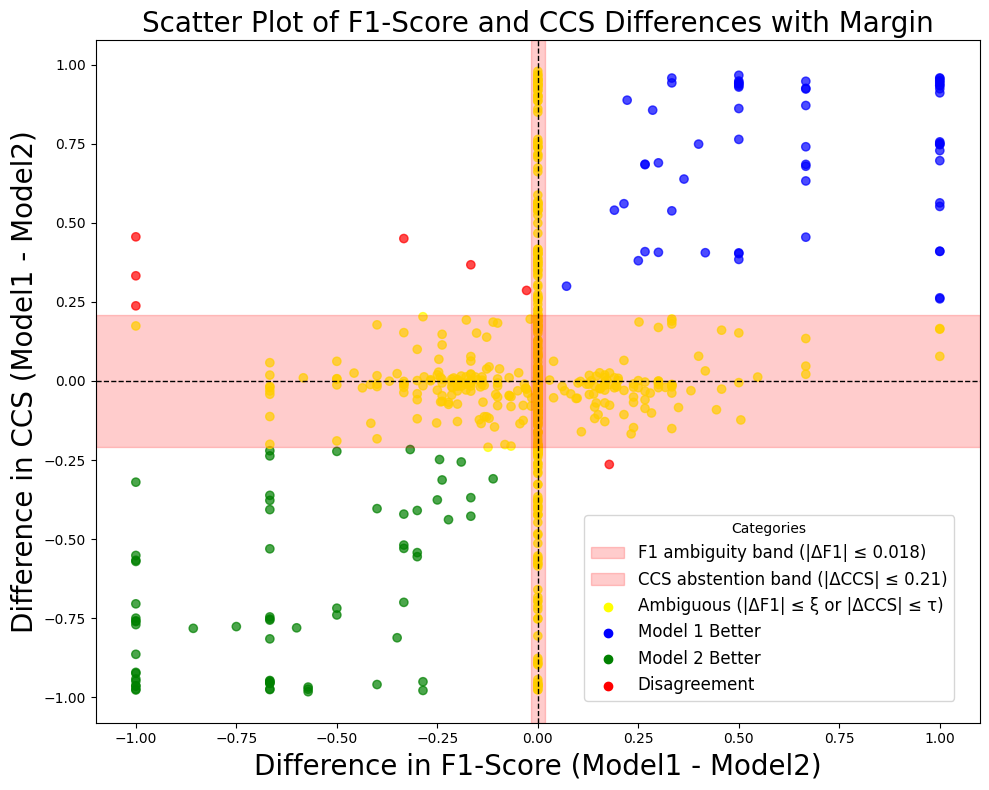}
		\caption{Comparison of F1-Score and CCS}
		\label{fig:f1_vs_ccs}
	\end{subfigure}
	\hfill
	\begin{subfigure}{0.32\textwidth}
		\centering
		\includegraphics[width=\textwidth]{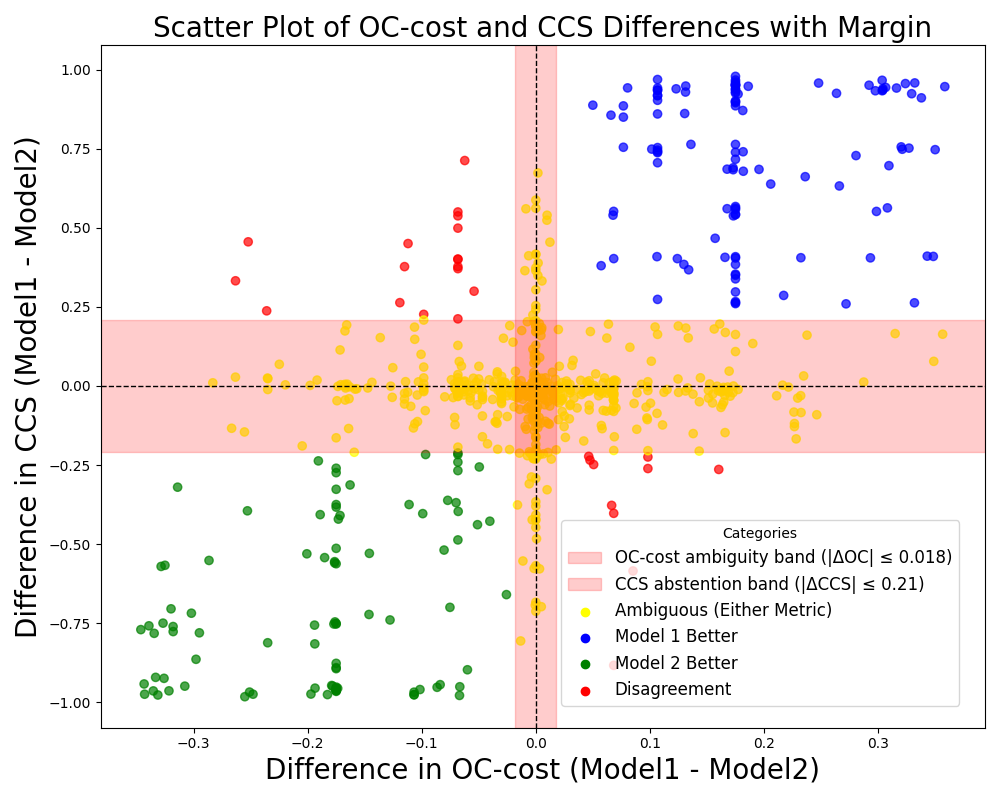}
		\caption{Comparison of OC-cost and CCS}
		\label{fig:oc_cost_vs_ccs}
	\end{subfigure}
    \hfill
    \begin{subfigure}{0.32\textwidth}
		\centering
		\includegraphics[width=\textwidth]{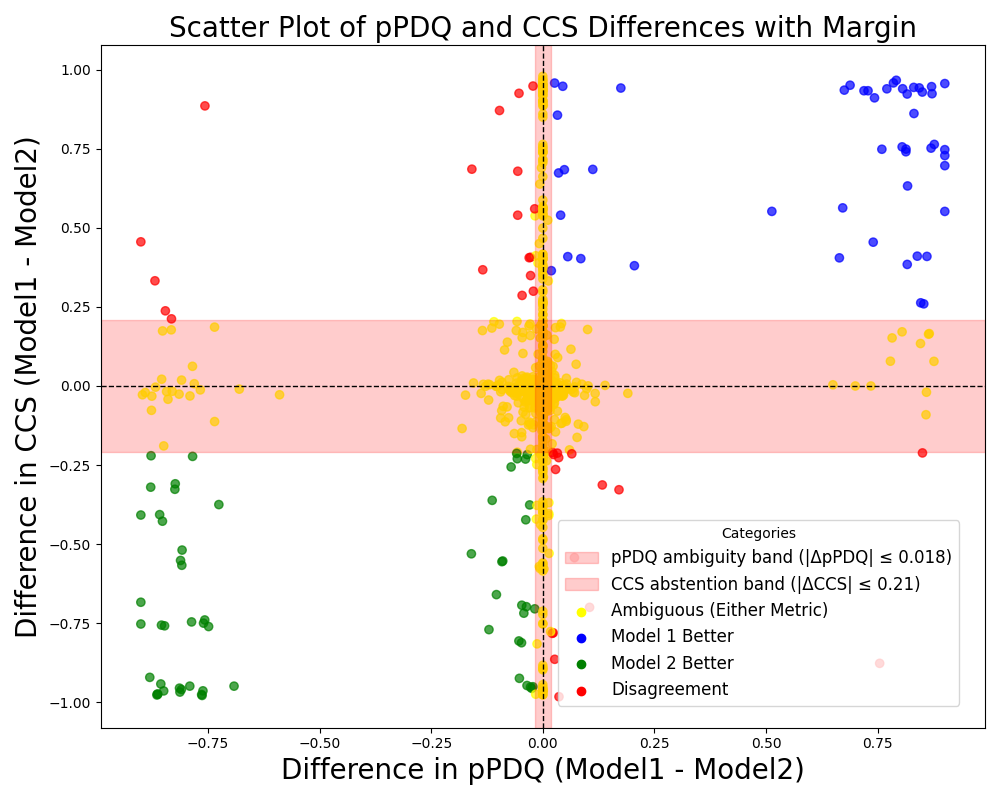}
		\caption{Comparison of pPDQ and CCS}
		\label{fig:ppdq_vs_ccs}
	\end{subfigure}
	\caption{Scatter plots comparing CCS with established metrics: F1-Score, pPDQ, and OC-cost.}
	\label{fig:ccs_comparison}
\end{figure*}
\begin{figure*}[!t]
	\centering
	\begin{subfigure}{0.32\textwidth}
		\centering
		\includegraphics[width=\textwidth]{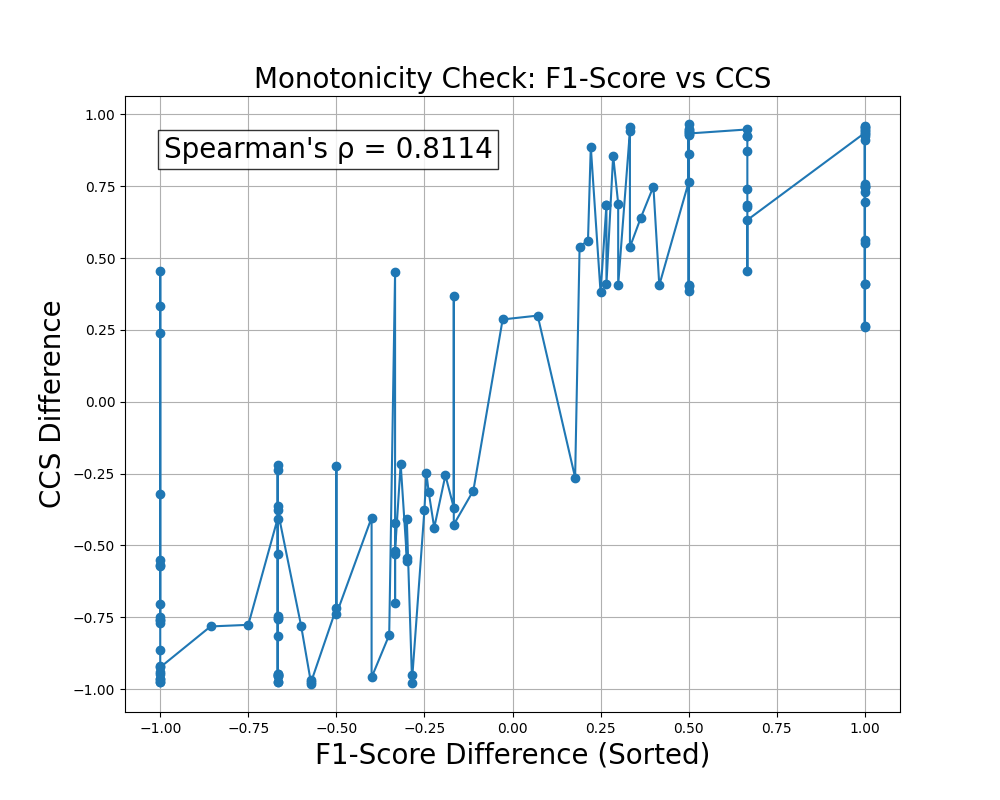}
		\caption{Sorted trend of F1-Score and CCS}
		\label{fig:f1_sorted_trend}
	\end{subfigure}
	\hfill
	\begin{subfigure}{0.32\textwidth}
		\centering
		\includegraphics[width=\textwidth]{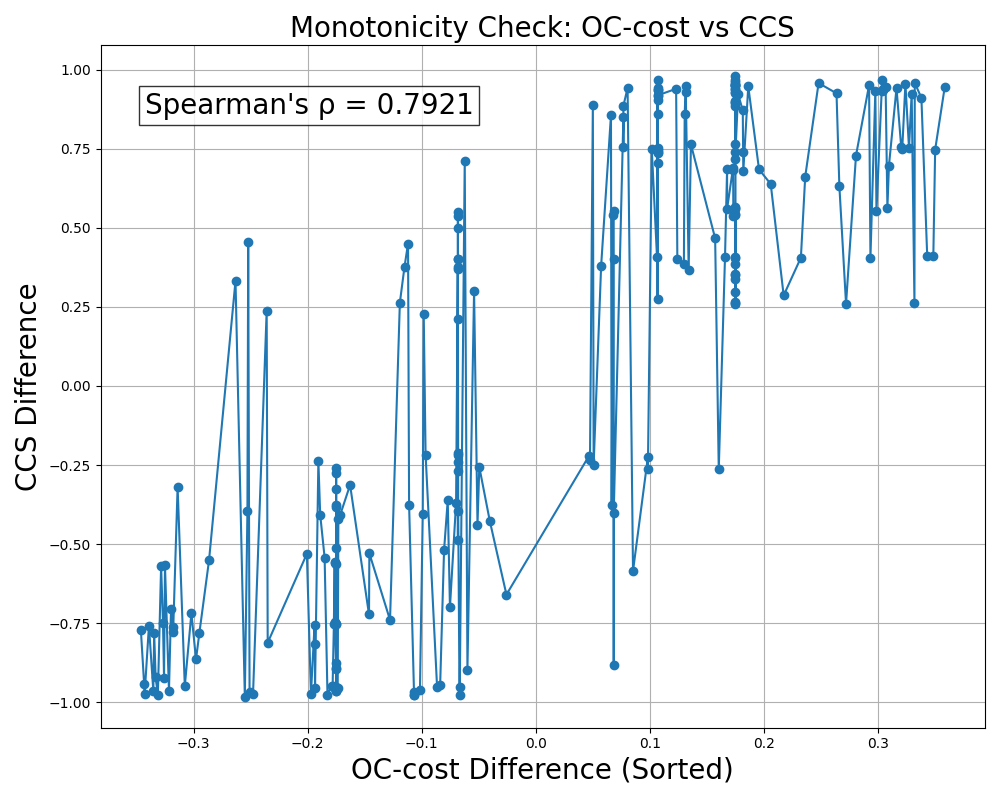}
		\caption{Sorted trend of OC-cost and CCS}
		\label{fig:oc_cost_sorted_trend}
	\end{subfigure}
    \hfill
    \begin{subfigure}{0.32\textwidth}
		\centering
		\includegraphics[width=\textwidth]{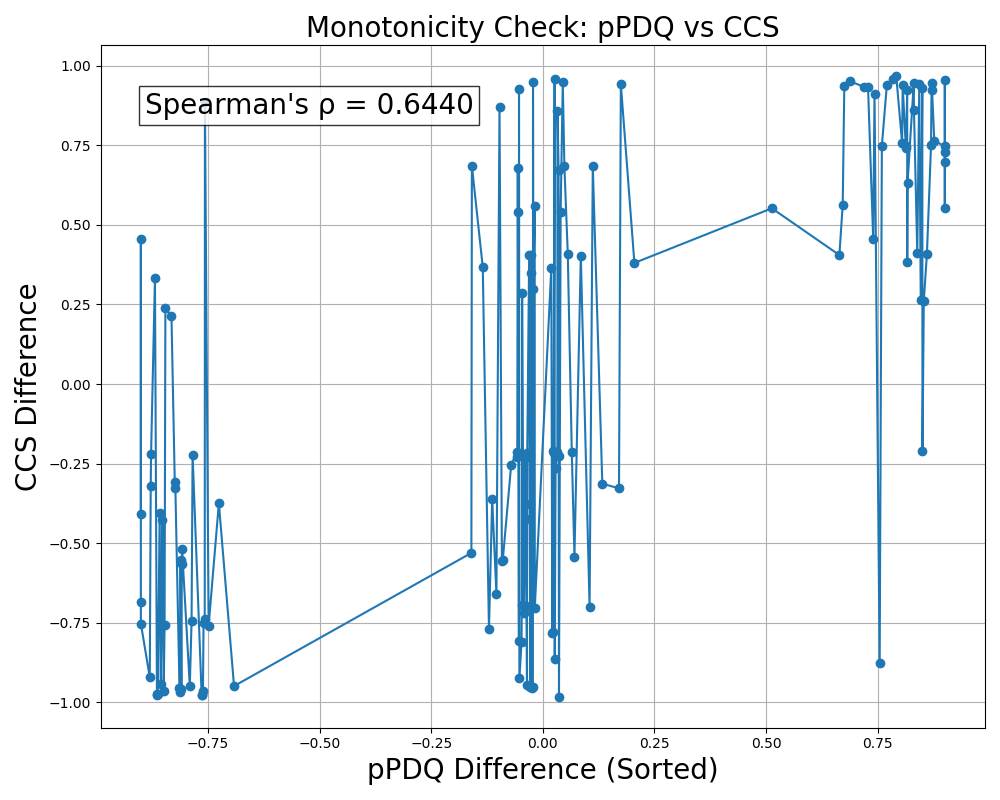}
		\caption{Sorted trend of pPDQ and CCS}
		\label{fig:ppdq_sorted_trend}
	\end{subfigure}
	\caption{Sorted trend analysis showing the alignment of CCS with F1-Score, pPDQ, and OC-cost.}
	\label{fig:sorted_trend_analysis}
    \vspace{-2mm}
\end{figure*}

\vspace{1mm}
\subsubsection{Assignment-based consensus computation}
After thresholding the IoU matrix $\Omega_{ij}$ to obtain $\Omega'_{ij}$, we determine correspondences between detections from the $i^{\text{th}}$ and $j^{\text{th}}$ augmentations using a one-to-one assignment formulation. This step resolves ambiguities that arise when multiple objects are present and when the numbers of detections differ across augmentations.

Let $\Omega'_{ij} \in \mathbb{R}^{N_i \times N_j}$ denote the filtered IoU matrix. To identify consistent detection pairs, we solve a linear assignment problem on $\Omega'_{ij}$ using the Hungarian algorithm. The assignment maximizes the total retained IoU while ensuring that each detection participates in at most one correspondence. Let $\mathcal{A}_{ij} \subseteq \{1,\dots,N_i\} \times \{1,\dots,N_j\}$ denote the set of matched index pairs returned by the Hungarian algorithm. The number of matched pairs is denoted by $|\mathcal{A}_{ij}|$.

We define the assignment matrix $\Pi_{ij} \in \mathbb{R}^{N_i \times N_j}$ as
\begin{equation}
\Pi_{ij}(k,l) =
\begin{cases}
\frac{1}{|\mathcal{A}_{ij}|}, & \text{if } (k,l) \in \mathcal{A}_{ij}, \\
0, & \text{otherwise}.
\end{cases}
\label{eq:pi_definition}
\end{equation}

By construction,
\begin{equation}
\sum_{k=1}^{N_i} \sum_{l=1}^{N_j} \Pi_{ij}(k,l) = 1.
\label{eq:pi_normalization_new}
\end{equation}

Using $\Pi_{ij}$, the pair-wise consensus score $\gamma_{ij}$ is defined as
\begin{equation}
\gamma_{ij}
\defeq
\sum_{k=1}^{N_i} \sum_{l=1}^{N_j}
\Pi_{ij}(k,l)\, \Omega'_{ij}(k,l).
\label{eq:assignment_consensus_score}
\end{equation}

Since $\Pi_{ij}$ distributes uniform mass over the matched pairs, the above expression simplifies to
\begin{equation}
\gamma_{ij}
=
\frac{1}{|\mathcal{A}_{ij}|}
\sum_{(k,l)\in\mathcal{A}_{ij}}
\Omega'_{ij}(k,l).
\label{eq:assignment_consensus_score_simplified}
\end{equation}

The case $|\mathcal{A}_{ij}| = 0$ arises when no detection pairs satisfy the IoU threshold, or when one of the augmentations yields no detections. In this situation, the set $\mathcal{A}_{ij}$ is empty, so the sum $\sum_{(k,l)\in\mathcal{A}_{ij}}\Omega'_{ij}(k,l)$ contains no terms and the normalization by $|\mathcal{A}_{ij}|$ would be undefined. This includes the case where both augmentations yield no detections, in which $\Omega'_{ij}$ does not exist, as well as the case where $\Omega'_{ij}$ is identically zero after thresholding; in both situations we define $\gamma_{ij} = 0$.

\vspace{1mm}
\subsubsection{Calculate CCS via aggregation}

After computing the pairwise consensus score $\gamma_{ij}$ for each ordered augmentation pair $(i,j)$ with $i \neq j$, we obtain the Cumulative Consensus Score (CCS) by averaging across all such pairs:

\begin{equation}
CCS
\defeq
\frac{1}{M(M-1)}
\sum_{i=1}^{M}
\sum_{\substack{j=1 \\ j \neq i}}^{M}
\gamma_{ij}.
\label{eq:cumulative_consensus_score}
\end{equation}

The normalization factor $M(M-1)$ accounts for all ordered augmentation pairs and ensures that CCS remains independent of the number of augmentations applied. By aggregating consensus across all pairs, CCS provides a normalized measure of localization consistency for the image that is independent of the number of augmentations applied.

\subsection{Minimal Link Between CCS and Detection Correctness}
\label{sec:theory_correctness_ccs}

In a highly simplified setting, we provide the following result to build intuition for why the TTDA agreement reflected by CCS can correlate with detection correctness.

\paragraph{Setting}
Assume a single object exists in the image. We apply $k$ augmentations.
For augmentation $t\in\{1,\dots,k\}$, let $X_t\in\{0,1\}$ indicate whether the detector is \emph{correct}.
If $X_t=1$, the detector outputs a perfect bounding box; otherwise no box is produced.
We consider two augmented views to be consistent if they yield the same outcome.

\paragraph{CCS under $i$ incorrect augmentations}
Let $i$ be the number of incorrect augmentations, hence $k-i$ are correct.
Among the $\binom{k}{2}$ unordered augmentation pairs, the number of consistent pairs equals
$\binom{k-i}{2}+\binom{i}{2}$.
Therefore, the CCS value conditioned on $i$ incorrect augmentations is
\begin{equation}
CCS_k(i)
=
\frac{1}{\binom{k}{2}}
\left(
\binom{k-i}{2}+\binom{i}{2}
\right).
\label{eq:ccs_given_i_incorrect}
\end{equation}

\paragraph{Expected CCS with correctness $p$}
Assuming i.i.d.\ trials $X_t\sim\mathrm{Bernoulli}(p)$, where $p$ denotes the detector's augmentation-wise correctness, the number of incorrect augmentations~$i$ follows a binomial distribution.
Thus, the expected CCS is
\begin{equation}
\small
\mathbb{E}[CCS_k]
=
\sum_{i=0}^{k}
\binom{k}{i}\, p^{k-i}(1-p)^i
\left[
\frac{1}{\binom{k}{2}}
\left(
\binom{k-i}{2}+\binom{i}{2}
\right)
\right].
\label{eq:expected_ccs_sum_form}
\end{equation}

\begin{lemma}
\label{thm:expected_ccs_gap}
Consider two object detectors $f_1$ and $f_2$ with augmentation-wise correctness $p_1$ and $p_2$, respectively,
under the idealized single-object setting in Sec.~\ref{sec:theory_correctness_ccs}.
Let $CCS_k$ be defined as in Eq.~\eqref{eq:ccs_given_i_incorrect}--Eq.~\eqref{eq:expected_ccs_sum_form}.
For any margin $\Omega\ge 0$, we have
\begin{equation}
\mathbb{E}[CCS_k^{f_1}]-\mathbb{E}[CCS_k^{f_2}]>\Omega
\label{eq:theorem_iff_left}
\end{equation}
\noindent if and only if
\begin{equation}
\scriptsize
\begin{aligned}
\sum_{i=0}^{k}\binom{k}{i}
\left(p_1^{k-i}(1-p_1)^i - p_2^{k-i}(1-p_2)^i\right)
\left[
\frac{1}{\binom{k}{2}}
\left(
\binom{k-i}{2}+\binom{i}{2}
\right)
\right]
>\Omega.
\end{aligned}
\label{eq:theorem_iff_expanded}
\end{equation}
\end{lemma}

\begin{proof} The creation of Eq.~\eqref{eq:theorem_iff_expanded} immediately follows by integrating  Eq.~\eqref{eq:expected_ccs_sum_form} into  Eq.~\eqref{eq:theorem_iff_left}. 
\end{proof}

The following result shows that, under simplifying theoretical assumptions, a more performant detector also has a higher CCS value.

\begin{lemma}[Monotonicity intuition]
\label{rem:monotonicity}
 For any $k\ge2$ and $1>p_1>p_2>0.5$, we have $\mathbb{E}[CCS_k^{f_1}] > \mathbb{E}[CCS_k^{f_2}]$. 
\end{lemma}

\begin{proof}
Consider $k\ge 2$ i.i.d.\ Bernoulli trials $X_1,\dots,X_k$ with $\Pr(X=1)=p$.
Let $i$ be the number of failures, so $i\sim \mathrm{Bin}(k,1-p)$.
The quantity
$CCS_k^{f}\;\defeq\;
\frac{\binom{k-i}{2}+\binom{i}{2}}{\binom{k}{2}}
$
is actually the fraction of unordered pairs $(a,b)$ with $a<b$ that have the same outcome
(both successes or both failures). Therefore, by linearity of expectation, its expectation equals
the probability that a single pair matches:
\begin{align}
\mathbb{E}\!\left[CCS_k^{f}\right]
&=\Pr(X_a=X_b) \nonumber\\
&=\Pr(X_a=1,X_b=1)+\Pr(X_a=0,X_b=0) \nonumber\\
&=p^2+(1-p)^2. \label{eq:toy_ECCS}
\end{align}
Hence

\vspace{-10mm}
\begin{align}
\mathbb{E}\!\left[CCS_k^{f_1}\right]\big|_{p=p_1}
-\mathbb{E}\!\left[CCS_k^{f_2}\right]\big|_{p=p_2}
\nonumber\\
=\Bigl(p_1^2+(1-p_1)^2\Bigr)
 -\Bigl(p_2^2+(1-p_2)^2\Bigr) \nonumber\\
=2(p_1-p_2)(p_1+p_2-1). \label{eq:toy_diff_factor}
\end{align}
If $1>p_1>p_2>\tfrac12$, then $p_1-p_2>0$ and $p_1+p_2-1>0$,
so
\[
\mathbb{E}\!\left[CCS_k^{f_1}\right]\big|_{p=p_1}
>
\mathbb{E}\!\left[CCS_k^{f_2}\right]\big|_{p=p_2}.
\]
\end{proof}

Note that our theoretical result is derived in an idealized setting and does not model localization noise, multiple objects, or class confusion. It therefore serves primarily to justify the monotonic intuition. While the proof is presented in this simplified regime for analytical clarity, our experiments in later sections show that the same ordering trend holds in realistic settings with noisy localization and multi-object assignment. In practice, localization noise makes IoU $< 1$ and multi-object scenes require assignment. Nevertheless, this relationship motivates CCS as a deployment signal. Our implementation adopts a conservative variant by setting $\gamma_{ij}=0$ when both augmentations yield no detections, so Lemma~\ref{rem:monotonicity} serves as intuition rather than an exact characterization of our implementation.

\section{Experiments and Results}
\label{sec:results}

\subsection{Experimental Setup}

\subsubsection{Dataset}
\label{sec:dataset}
We train object detectors on Open Images Dataset (OID)~\cite{kuznetsova2020open,Krasin2017openimages} and evaluate on KITTI~\cite{geiger2012ready,geiger2013vision}.

\subsubsection{Evaluation metrics}
\label{sec:Evaluation_metric}
We compare CCS against F1-score, OC-cost, and pPDQ.

\textbf{F1-score:} Following~\cite{Le2018}, we apply one-to-one matching at IoU threshold $\alpha_{\text{iou}}$. A prediction is a TP if its class matches the ground truth and IoU $\ge \alpha_{\text{iou}}$, otherwise FP; unmatched ground-truth boxes are FN. The F1-score is
\begin{equation}
\text{F1} = 2\,\frac{\text{Precision}\cdot\text{Recall}}{\text{Precision}+\text{Recall}}.
\label{eq:f1score}
\end{equation}

\textbf{OC-cost:} OC-cost~\cite{Otani_2022_CVPR} evaluates detectors via minimum-cost bipartite matching between predictions and ground truths. Unmatched boxes incur a constant penalty $\beta_{\text{dummy}}$ (set to $0.6$, which aligns with human preference as recommended in~\cite{Otani_2022_CVPR}). The per pair matching cost is
\begin{equation}
C_{\text{total}}=\lambda C_{\text{loc}}+(1-\lambda) C_{\text{cls}},
\label{eq:oc_cost}
\end{equation}
where $C_{\text{loc}}=1-\mathrm{IoU}$. 
In our setup, we set $\lambda=1$ to focus exclusively on localization.

\textbf{pPDQ}: pPDQ~\cite{hall2020probabilistic} combines spatial and semantic uncertainty via the geometric mean between a ground truth~$\mathcal{G}_i^{\text{img}}$ and a detection $\mathcal{D}_j^{\text{img}}$:
\begin{equation}
\text{pPDQ}(\mathcal{G}_i^{\text{img}},\mathcal{D}_j^{\text{img}})
=\sqrt{Q_S(\mathcal{G}_i^{\text{img}},\mathcal{D}_j^{\text{img}})
\cdot Q_L(\mathcal{G}_i^{\text{img}},\mathcal{D}_j^{\text{img}})}.
\label{eq:ppdq}
\end{equation}
Here $Q_S$ measures how well the spatial probability distribution of a detection aligns with the ground truth, while $Q_L$ is the probability the detector assigns to the correct class label. 
Unlike a simple $L_2$ norm, $Q_S$ is formally defined as
\begin{equation}
\begin{split}
Q_S(\mathcal{G}_i^{\text{img}},\mathcal{D}_j^{\text{img}})
= \exp\!\Big(-\big[&
L_{\text{FG}}(\mathcal{G}_i^{\text{img}},\mathcal{D}_j^{\text{img}}) \\
&+ L_{\text{BG}}(\mathcal{G}_i^{\text{img}},\mathcal{D}_j^{\text{img}})\big]\Big).
\end{split}
\end{equation}

where $L_{\text{FG}}$ is the average negative log-probability assigned to foreground pixels and $L_{\text{BG}}$ penalizes probability mass assigned outside the ground-truth bounding box. 

\noindent\textit{Implementation note.} Although PDQ was originally defined for pixel-accurate masks~\cite{hall2020probabilistic}, we adapt it to bounding boxes by assigning probability $1-\epsilon$ inside the predicted box and $\epsilon$ outside, with $\epsilon=0.1$ to enforce strict localization penalties. As the evaluated detectors here (Faster R-CNN, RetinaNet, SSD) are deterministic and do not output explicit spatial uncertainty distributions, this adaptation (with fixed $\epsilon$) should be interpreted as a standardized approximation for consistent supervised comparison across deterministic architectures. By this, PDQ serves as a supervised baseline metric rather than a full probabilistic evaluation. 

\begin{table}[t]
\centering
\caption{Comparison of old and new object detectors trained on 2000 Car and 2000 Truck samples from OID~\cite{Krasin2017openimages}.}
\label{tab:model_comparison}
\footnotesize
\setlength{\tabcolsep}{3pt}
\renewcommand{\arraystretch}{1.1}
\begin{tabular}{|c|c|c|}
\hline
\textbf{Metric} &
\makecell{\textbf{Model 1:} FRCNN \\ (50 ep.)} &
\makecell{\textbf{Model 2:} RetinaNet \\ (100 ep.)} \\
\hline
COCO mAP (baseline) & 0.384 & 0.409 \\ \hline
mAP (after training) & 0.4307 & 0.4979 \\ \hline
Validation Loss & 0.1872 & 0.1746 \\ \hline
\end{tabular}
\vspace{-5mm}
\end{table}

\subsection{Congruence of CCS with Established Metrics}
This subsection evaluates how well CCS aligns with standard supervised metrics (Sec.~\ref{sec:Evaluation_metric}).
In Fig.~\ref{fig:ccs.computation.workflow}, each image is processed by two object detectors $f_{1}$ and $f_{2}$ (Tab.~\ref{tab:model_comparison}).
For each supervised metric $M \in \{\mathrm{F1},\,\mathrm{pPDQ},\,\mathrm{OC\text{-}Cost}\}$, we obtain per-image scores $M_{f_1}$ and $M_{f_2}$ and form the deltas
\begin{equation}
\Delta M = M_{f_{1}} - M_{f_{2}}.
\label{eq:deltas}
\end{equation}
Similarly, the difference of CCS between the two models is
\begin{equation}
\Delta CCS = CCS_{f_{1}} - CCS_{f_{2}}.
\label{eq:delta_ccs}
\end{equation}

Fig.~\ref{fig:ccs_comparison} plots $\Delta CCS$ against $\Delta M$ for the model pairs under study.
To avoid attributing meaning to near-ties, we introduce (i) an \emph{ambiguity band} for the supervised reference, and (ii) a \emph{CCS abstention band} around zero.

\paragraph{F1 ambiguity band ($\xi$)}
We use F1 as the reference to determine whether supervised evaluation is image-wise conclusive.
For a fixed model pair and test set, we compute per-image $\Delta \mathrm{F1}$ values (Eq.~\eqref{eq:deltas}).
We then estimate the uncertainty of the \emph{mean} $\Delta \mathrm{F1}$ via nonparametric bootstrap and define $\xi$ as the half-width of the resulting $95\%$ confidence interval (CI).
Images with $|\Delta \mathrm{F1}| \le \xi$ are treated as \emph{F1-ambiguous}, while images with $|\Delta \mathrm{F1}| > \xi$ are treated as \emph{F1-unambiguous}.

\paragraph{Calibrating CCS abstention threshold ($\tau$) from F1-ambiguous images}
We calibrate $\tau$ using the F1-ambiguous subset so that CCS abstains on a controlled fraction of cases where the supervised reference is itself inconclusive.
Let $\mathcal{A}_{\mathrm{F1}}=\{\text{images}:|\Delta \mathrm{F1}|\le\xi\}$.
We set
\begin{equation}
\tau(p) \;=\; \operatorname{Percentile}_p\!\big(\{\,|\Delta CCS| : \text{image}\in\mathcal{A}_{\mathrm{F1}}\,\}\big),
\label{eq:tau_percentile}
\end{equation}
and report a sweep over $p\in\{50,75,90,95\}$.
A larger $p$ yields a more conservative $\tau$ (higher abstention), while a smaller $p$ yields a less conservative $\tau$ (higher coverage).

\paragraph{Kept set definition}
For a given $(\xi,\tau)$, we restrict analysis to images where \emph{both} the supervised reference and CCS express a confident preference:
\begin{equation}
\mathcal{K}(\xi,\tau) \;=\; \{\text{images}: |\Delta \mathrm{F1}|>\xi \;\wedge\; |\Delta CCS|>\tau\}.
\label{eq:kept_set}
\end{equation}
Images outside $\mathcal{K}(\xi,\tau)$ are treated as abstentions, either because the supervised reference is inconclusive (F1-ambiguous) or because CCS assigns a near-zero difference.

\begin{table}[t]
\centering
\caption{Effect of percentile $p$ on CCS abstention and alignment for the illustrative model pair (KITTI test set).}
\label{tab:tau_sweep}
\footnotesize
\setlength{\tabcolsep}{3pt}
\renewcommand{\arraystretch}{1.1}
\begin{tabularx}{\columnwidth}{|c|c|Y|Y|c|c|}
\hline
\textbf{$p$ (\%)} & $\boldsymbol{\tau}$ &
\textbf{Keep (unamb.)} &
\textbf{Abstain (amb.)} &
$\boldsymbol{\rho}$ &
\textbf{Congr.} \\
\hline
50  & 0.0287 & 0.6731 & 0.5008 & 0.7585 & 0.7510 \\
\textbf{75} & 0.2134 & 0.3626 & 0.7504 & 0.7933 & 0.9470 \\
90  & 0.6652 & 0.2060 & 0.8995 & 0.7398 & 1.0000 \\
95  & 0.9059 & 0.1154 & 0.9498 & 0.7090 & 1.0000 \\
\hline
\end{tabularx}
\vspace{-5mm}
\end{table}

\paragraph{Percentile selection for the illustrative model pair}
Tab.~\ref{tab:tau_sweep} summarizes the effect of varying the percentile $p$ used to define $\tau$, where $\tau$ is computed from the distribution of $|\Delta CCS|$ restricted to the F1-ambiguous subset determined by $\xi$ (i.e., $|\Delta \mathrm{F1}| \le \xi$). 
Thus, the sweep is performed with respect to the $\Delta$F1-based calibration.

Lower percentiles (e.g., $p{=}50$) are permissive, retaining many unambiguous cases but abstaining on only half of the ambiguous subset, which reduces congruence and ranking consistency. 
Higher percentiles ($p{=}90$ and $95$) enforce strict abstention, eliminating nearly all ambiguous cases and achieving perfect sign agreement, but at the cost of retaining too few samples and lowering Spearman’s $\rho$.

The choice $p{=}75$ provides the best trade-off: it abstains on approximately $75\%$ of ambiguous cases while retaining a sufficient number of unambiguous images for reliable correlation analysis, and yields the highest observed Spearman correlation together with high congruence. 
The resulting $\tau$ is then fixed and used consistently when comparing CCS against $\Delta$OC-Cost and $\Delta$pPDQ.

\subsection{Observations from Fig.~\ref{fig:ccs_comparison}}
In Fig.~\ref{fig:ccs_comparison}, each point corresponds to one test image.
Colors encode agreement between $\Delta M$ and $\Delta CCS$, with an explicit abstention region:

\begin{itemize}
  \item \textbf{Yellow (abstain):} at least one measure abstains, i.e.,
  $|\Delta \mathrm{F1}|\le\xi$ \emph{or} $|\Delta CCS|\le\tau$.
  This includes (a) supervised near-ties (F1-ambiguous images) and (b) cases where CCS itself assigns near-zero difference and therefore does not support a confident ranking.
  \item \textbf{Blue (agree, $f_1$ better):} $\Delta M>0$ and $\Delta CCS>0$ on the kept set $\mathcal{K}(\xi,\tau)$.
  \item \textbf{Green (agree, $f_2$ better):} $\Delta M<0$ and $\Delta CCS<0$ on the kept set $\mathcal{K}(\xi,\tau)$.
  \item \textbf{Red (disagree):} $\operatorname{sign}(\Delta M)\neq \operatorname{sign}(\Delta CCS)$ on the kept set $\mathcal{K}(\xi,\tau)$.
\end{itemize}

Thus, blue/green/red points correspond to images for which both the supervised reference and CCS are confident, while yellow points are deliberately excluded from congruence and Spearman’s $\rho$ computations to avoid over-interpreting near-ties.

\begin{table}[t]
\centering
\caption{Alignment of CCS with supervised metrics for the illustrative model pair in Tab.~\ref{tab:model_comparison}.
All results are computed over 1000 test images (KITTI test set).
Calibration uses $\Delta$F1 ($\xi=0.018$, $\tau=0.21$ at the 75th percentile).}
\label{tab:ccs_results}
\footnotesize
\setlength{\tabcolsep}{2pt}
\renewcommand{\arraystretch}{1.05}
\begin{tabularx}{\columnwidth}{|
>{\raggedright\arraybackslash}p{0.30\columnwidth}
|>{\centering\arraybackslash}p{0.07\columnwidth}
|>{\centering\arraybackslash}p{0.07\columnwidth}
|>{\centering\arraybackslash}p{0.06\columnwidth}
|>{\centering\arraybackslash}p{0.06\columnwidth}
|>{\centering\arraybackslash}p{0.06\columnwidth}
|>{\centering\arraybackslash}p{0.14\columnwidth}
|>{\centering\arraybackslash}p{0.10\columnwidth}|
}
\hline
\textbf{Metric} &
\textbf{Yel.} &
\textbf{Kept} &
\textbf{Grn.} &
\textbf{Blu.} &
\textbf{Red} &
\textbf{Congr.}\textbf{(\%)} &
\textbf{$\rho$}
\\ \hline
$\Delta$F1 vs $\Delta$CCS
& 870 & 130 & 65 & 59 & 6 & 95.38 & 0.8114 \\ \hline
$\Delta$OC-cost vs $\Delta$CCS
& 766 & 234 & 94 & 114 & 26 & 88.89 & 0.7921 \\ \hline
$\Delta$pPDQ vs $\Delta$CCS
& 860 & 140 & 58 & 49 & 33 & 76.43 & 0.6440 \\ \hline
\end{tabularx}
\vspace{-5mm}
\end{table}

\subsection{Sorted Trend Analysis for Congruence Assessment}

Fig.~\ref{fig:ccs_comparison} presents scatter plots of $\Delta CCS$ against $\Delta M$ for $M \in \{\mathrm{F1},\,\mathrm{pPDQ},\,\mathrm{OC\text{-}Cost}\}$. 
Each point corresponds to one test image, and agreement categories are determined on the kept set $\mathcal{K}(\xi,\tau)$ defined in Eq.~\eqref{eq:kept_set}.

To assess whether CCS preserves the ordering induced by a supervised metric, we perform a sorted trend analysis (Fig.~\ref{fig:sorted_trend_analysis}). 
For each metric $M$, images in $\mathcal{K}(\xi,\tau)$ are sorted according to $\Delta M$, and the corresponding $\Delta CCS$ values are plotted in the same order. 
If CCS is consistent with $M$, the resulting curve exhibits a monotonic trend: images strongly favoring $f_1$ under $M$ (large positive $\Delta M$) tend to have positive $\Delta CCS$, and images favoring $f_2$ tend to have negative $\Delta CCS$. Agreement on $\mathcal{K}(\xi,\tau)$ is quantified using two complementary measures. 
First, directional agreement (congruence) is defined as 
$\mathbb{P}\big[\operatorname{sign}(\Delta M)=\operatorname{sign}(\Delta CCS)\big]$, 
which, in empirical form, equals $(\text{Green} + \text{Blue}) / \text{Kept} \times 100\%$. 
Second, we compute Spearman’s rank correlation coefficient $\rho(\Delta M,\Delta CCS)$, which evaluates monotonic consistency between the two quantities.

Spearman’s $\rho$ is used instead of a simple vector difference or Pearson correlation because the objective is to assess ranking consistency rather than absolute scale agreement. 
Different metrics may vary in magnitude or exhibit nonlinear scaling while preserving the same ordering, which Spearman’s $\rho$ captures directly.

\subsection{Inference from Tab.~\ref{tab:ccs_results}}
Tab.~\ref{tab:ccs_results} analyzes the illustrative model pair in Tab.~\ref{tab:model_comparison} using calibration parameters $\xi$ and $\tau$ derived from $\Delta$F1. A substantial portion of images lies in the abstention region, reflecting the similar supervised performance of the two detectors and the prevalence of image-level near-ties captured by the ambiguity band $\xi$. On the kept set (Eq.~\eqref{eq:kept_set}), CCS exhibits strong directional agreement with $\Delta$F1 (95.38\%) and high monotonic consistency ($\rho = 0.8114$). Under the same calibration, alignment with $\Delta$OC-cost and $\Delta$pPDQ remains substantial (88.89\% and 76.43\%; $\rho = 0.7921$ and $0.6440$); the weaker pPDQ alignment is expected, as pPDQ jointly penalizes localization and classification confidence, a dimension CCS does not capture.

\begin{table*}[t]
\centering
\caption{Alignment of CCS with supervised metrics across diverse model comparison scenarios.
All results are computed over 1000 test images from KITTI.
For each model pair, \boldmath$\boldsymbol{\xi}$\unboldmath\ (bootstrap CI half-width of mean $\Delta$F1) and
\boldmath$\boldsymbol{\tau}$\unboldmath\ (75th percentile of $|\Delta\mathrm{CCS}|$ over F1-ambiguous images) are shown in the model column.
Congruence and Spearman's $\rho$ are computed on the kept set.}
\label{tab:ccs_results_2}
\footnotesize
\setlength{\tabcolsep}{2pt}
\renewcommand{\arraystretch}{1.9}
\begin{tabularx}{\textwidth}{|
>{\centering\arraybackslash}p{0.19\textwidth}
|>{\centering\arraybackslash}p{0.30\textwidth}
|>{\centering\arraybackslash}p{0.08\textwidth}
|>{\centering\arraybackslash}p{0.045\textwidth}
|>{\centering\arraybackslash}p{0.045\textwidth}
|>{\centering\arraybackslash}p{0.045\textwidth}
|>{\centering\arraybackslash}p{0.045\textwidth}
|>{\centering\arraybackslash}p{0.045\textwidth}
|>{\centering\arraybackslash}p{0.06\textwidth}
|>{\centering\arraybackslash}p{0.06\textwidth}|
}
\hline
\textbf{Training Setup} &
\textbf{Model Pair} &
\textbf{$\Delta M$} &
\textbf{Yel.} &
\textbf{Kept} &
\textbf{Grn.} &
\textbf{Blu.} &
\textbf{Red} &
\textbf{Congr.} &
\textbf{$\rho$}
\\ \hline
\multirow{3}{*}{\makecell[c]{
\textbf{Cross-Architecture}\\
Images/class: 6000\\
Labels: Car, Truck
}} &
\multirow{3}{*}{\parbox[c][5.2\baselineskip][c]{0.30\textwidth}{\centering
Model1: Faster R-CNN ResNet-101\\
2 epochs, mAP: 0.3308\\
Model2: SSD300 VGG16\\
250 epochs, mAP: 0.5053\\
\fbox{\footnotesize\textbf{$\xi=0.016,\ \tau=0.119$}}
}} &
$\Delta$F1
& 852 & 148 & 81 & 54 & 13 & 91.22 & 0.8085
\\ \cline{3-10}
& & $\Delta$OC-cost
& 742 & 258 & 140 & 90 & 28 & 89.15 & 0.7482
\\ \cline{3-10}
& & $\Delta$pPDQ
& 825 & 175 & 71 & 71 & 33 & 81.14 & 0.7250
\\ \hline
\multirow{3}{*}{\makecell[c]{
\textbf{Same Architecture}\\
Old: 2000 imgs/class\\
New: 6000 imgs/class
}} &
\multirow{3}{*}{\parbox[c][5.2\baselineskip][c]{0.30\textwidth}{\centering
Model1: RetinaNet ResNet-50\\
50 epochs, mAP: 0.4837\\
Model2: RetinaNet ResNet-50\\
150 epochs, mAP: 0.5194\\
\fbox{\footnotesize\textbf{$\xi=0.016,\ \tau=0.111$}}
}} &
$\Delta$F1
& 872 & 128 & 59 & 55 & 14 & 89.06 & 0.8264
\\ \cline{3-10}
& & $\Delta$OC-cost
& 780 & 220 & 115 & 86 & 19 & 91.36 & 0.7676
\\ \cline{3-10}
& & $\Delta$pPDQ
& 855 & 145 & 52 & 51 & 42 & 71.03 & 0.4984
\\ \hline
\multirow{3}{*}{\makecell[c]{
\textbf{Same Architecture}\\
Weak vs Strong\\
Old: 1 epoch, 2000/class\\
New: 50 epochs, 6000/class
}} &
\multirow{3}{*}{\parbox[c][5.2\baselineskip][c]{0.30\textwidth}{\centering
Model1: Faster R-CNN ResNet-50\\
1 epoch, mAP: 0.4472\\
Model2: Faster R-CNN ResNet-50\\
50 epochs, mAP: 0.5098\\
\fbox{\footnotesize\textbf{$\xi=0.017,\ \tau=0.062$}}
}} &
$\Delta$F1
& 783 & 217 & 113 & 46 & 58 & 73.27 & 0.7031
\\ \cline{3-10}
& & $\Delta$OC-cost
& 740 & 260 & 140 & 60 & 60 & 76.92 & 0.6662
\\ \cline{3-10}
& & $\Delta$pPDQ
& 772 & 228 & 112 & 30 & 86 & 62.28 & 0.4049
\\ \hline
\end{tabularx}
\vspace{-5mm}
\end{table*}

\subsection{Inference from Tab.~\ref{tab:ccs_results_2}}
Tab.~\ref{tab:ccs_results_2} extends the analysis to heterogeneous comparison scenarios. For each model pair, the ambiguity band $\xi$ is computed via bootstrap confidence intervals on the mean $\Delta$F1. The CCS abstention threshold $\tau$ is then derived from the distribution of $|\Delta\mathrm{CCS}|$ within the F1-ambiguous subset. While the percentile used to compute $\tau$ (75th percentile in our experiments) controls the strictness of abstention, the calibration procedure itself remains data-driven.

Across all scenarios, CCS maintains strong directional agreement with supervised metrics, particularly for $\Delta$F1 and $\Delta$OC-cost, and consistently high Spearman correlations. In settings where the imbalance between green and blue counts becomes pronounced, Spearman's $\rho$ decreases, as it measures rank consistency rather than mere sign agreement. Nevertheless, the overall congruence remains high, indicating that CCS continues to identify the correct direction of performance difference even when magnitude ordering becomes less strictly monotonic.

Taken together, the results from both the illustrative example (Tab.~\ref{tab:ccs_results}) and the heterogeneous comparison scenarios (Tab.~\ref{tab:ccs_results_2}) show that whenever supervised evaluation yields a decisive preference, CCS preserves both the direction and the relative ordering of model performance differences. This consistency across architectures, training scales, and performance gaps supports the role of CCS as a model-agnostic, label-free monitoring signal.

\subsection{Comparison with Deployment-Time Output Signals}

To assess whether simpler label-free heuristics could substitute CCS, we compare CCS against scalar signals derived directly from detector outputs without ground-truth annotations. We evaluate three commonly used proxy indicators:
(i) \emph{mean detection confidence} \cite{guo2017calibration,hendrycks2017baseline};
(ii) \emph{detection count stability}, defined as the standard deviation of the number of detections across augmentations for each image \cite{gama2014survey};
and (iii) \emph{naïve IoU consistency}, measuring average spatial overlap agreement across augmentations without structured object association \cite{hendrycks2019robustness}.

For a model pair $(f_1,f_2)$, we compute image-level differences $\Delta H = H_{f_1} - H_{f_2}$ and compare them against $\Delta$F1 using the same ranking protocol as CCS. 
Since these signals have different numeric scales, we compute a signal-specific abstention threshold $\tau_H$ for each metric, defined as the 75th percentile of $|\Delta H|$ over the F1-ambiguous subset ($|\Delta \mathrm{F1}| \le \xi$). 
Agreement is evaluated on the kept set given by ($|\Delta \mathrm{F1}| > \xi \;\wedge\; |\Delta H| > \tau_H$).

Table~\ref{tab:label_free_comparison} reports results for the illustrative model pair of Tab.~\ref{tab:model_comparison}. 
Despite scale-aware calibration, all heuristics exhibit negligible monotonic alignment with supervised ranking ($|\rho| < 0.1$) and directional agreement close to random consistency. 
Over 90\% of images are classified as inconclusive under this protocol, resulting in small kept subsets.

In contrast, CCS (Tab.~\ref{tab:ccs_results}) achieves substantially stronger monotonic and directional alignment under identical calibration. This pattern holds across the additional model-comparison scenarios in Tab.~\ref{tab:ccs_results_2}, where heuristic signals consistently exhibit weaker correlation relative to CCS.

\begin{table}[t]
\centering
\caption{Alignment of deployment-time heuristic signals with supervised ranking ($\Delta$F1) for the illustrative model pair (1000 KITTI test images). Calibration uses $\xi=0.0181$. For each heuristic $H$, a signal-specific threshold $\tau_H$ is computed.}
\label{tab:label_free_comparison}
\footnotesize
\setlength{\tabcolsep}{3pt}
\renewcommand{\arraystretch}{1.1}
\begin{tabularx}{\columnwidth}{|
>{\raggedright\arraybackslash}X
|>{\centering\arraybackslash}p{0.10\columnwidth}
|>{\centering\arraybackslash}p{0.14\columnwidth}
|>{\centering\arraybackslash}p{0.10\columnwidth}
|>{\centering\arraybackslash}p{0.15\columnwidth}
|>{\centering\arraybackslash}p{0.10\columnwidth}|
}
\hline
\textbf{Signal} & $\boldsymbol{\rho}$ & \textbf{Congr.(\%)} & \textbf{Kept} & \textbf{Inconcl.(\%)} & $\boldsymbol{\tau_H}$ \\ \hline
Mean Confidence & $-0.09$ & $55.41$ & $74$ & $92.61$ & $1.00$ \\ \hline
Detection Count Stability & $0.07$ & $46.32$ & $95$ & $90.51$ & $1.50$ \\ \hline
Na\"ive IoU Consistency & $0.01$ & $56.00$ & $100$ & $90.01$ & $0.52$ \\ \hline
\end{tabularx}

\vspace{-5mm}
\end{table}

\subsection{Robustness to Augmentation Seeds, Architectures, and Datasets}
\label{sec:robustness}
Since CCS relies on test-time data augmentation (TTDA), we verify that its alignment with supervised metrics is not sensitive to the choice of augmentation seed. Tab.~\ref{tab:seed_robustness} reports representative results for five seeds under the illustrative detector setup of Tab.~\ref{tab:model_comparison} (15 seeds were evaluated in total). Spearman’s $\rho$ values remain tightly clustered: for $\Delta$F1 vs.\ $\Delta$CCS we obtain 
$\rho = 0.8205 \pm 0.0063$, 
for $\Delta$OC-cost vs.\ $\Delta$CCS, 
$\rho = 0.7869 \pm 0.0046$, 
and for $\Delta$pPDQ vs.\ $\Delta$CCS, 
$\rho = 0.6325 \pm 0.0080$. 
The small standard deviations indicate stable ranking consistency across seeds. Robustness is further supported by consistent behavior across architectures and training regimes (Tab.~\ref{tab:ccs_results_2}).

\paragraph{TTDA configuration}
We employ nine mild, non-geometric photometric transformations per image (\emph{brightness, contrast, blur, noise, sharpen, color shift}), with parameters sampled from narrow, bounded intervals (e.g., brightness $\alpha \in [0.9,1.1]$, $\beta \in [-10,10]$; Gaussian noise $\sigma \in [0.005,0.01]$; blur kernel $\in \{3,5\}$). 
These augmentations correspond to common corruption families used in robustness benchmarks which emulate realistic sensor or illumination variation rather than adversarial distortions\cite{hendrycks2019robustness,michaelis2019robustdet} and preserve semantic labels as recommended in augmentation surveys \cite{shorten2019survey,mumuni2022survey}. 

Geometric transformations are excluded so that CCS evaluates stability under appearance variation rather than systematic spatial displacement. As augmentation parameters are drawn from fixed, narrow ranges, changing the seed alters only the sampled value within the interval, explaining the observed stability of $\rho$.

\paragraph{Dataset robustness}
We additionally repeated the analysis on the COCO validation split and on BDD100K using the same model comparison configurations and calibration procedure. The qualitative behavior remained consistent: CCS preserved directional agreement with supervised metrics across architectures and training regimes, and the percentile-based abstention mechanism exhibited similar coverage–reliability trade-offs. Detailed COCO and BDD100K tables are omitted for brevity but mirror the KITTI trends.

\begin{table}[t]
\centering
\caption{Robustness of CCS alignment across different augmentation seeds for the illustrative model pair (Tab. \ref{tab:model_comparison}).}
\label{tab:seed_robustness}
\renewcommand{\arraystretch}{1.2}
\setlength{\tabcolsep}{4pt}
\footnotesize
\resizebox{\linewidth}{!}{
\begin{tabular}{|c|c|c|c|}
\hline
\textbf{Aug. Seed} & 
$\rho(\Delta$F1, $\Delta$CCS) & 
$\rho(\Delta$OC-cost, $\Delta$CCS) & 
$\rho(\Delta$pPDQ, $\Delta$CCS) \\
\hline
15  & 0.8170 & 0.7885 & 0.6326 \\
33  & 0.8263 & 0.7946 & 0.6447 \\
55  & 0.8219 & 0.7832 & 0.6221 \\
101 & 0.8107 & 0.7821 & 0.6316 \\
150 & 0.8265 & 0.7865 & 0.6316 \\
\hline
\end{tabular}
}
\end{table}

\subsection{Platform and Runtime}

\begin{table}[t]
\centering
\footnotesize
\caption{CCS post-processing time per image on CPU (1000 KITTI test images; $M{=}9$ augmentations).}
\label{tab:ccs_runtime}
\setlength{\tabcolsep}{3pt}
\renewcommand{\arraystretch}{1.05}
\begin{tabular}{|l|c|}
\hline
\textbf{Host CPU} & 2$\times$ EPYC 7742 (256 threads) \\ \hline
\textbf{Ordered pairs $M(M{-}1)$} & 72 \\ \hline
\textbf{Boxes/aug ($N_i$)} & $\le 5$ \\ \hline
\textbf{Per–pair time [ms]} & median $\approx 0.05$ \\ \hline
\textbf{Per–image time [ms]} & 0.09–32.53 (median $\approx 3.9$) \\ \hline
\end{tabular}
\vspace{-3mm}
\end{table}

Experiments were conducted on a server with 8$\times$ NVIDIA A100 GPUs and dual-socket AMD EPYC 7742 CPUs. Detector inference runs on GPU and CCS post-processing on CPU. With $M{=}9$ augmentations evaluated sequentially on one A100, inference takes $26$–$33$\,ms per augmentation (median $\approx 265$\,ms per image), while CCS adds a median overhead of $3.9$\,ms per image (Tab.~\ref{tab:ccs_runtime}). For each ordered pair $(i,j)$, CCS computes IoU, thresholds, and performs Hungarian matching. The per-image complexity is
\[
\mathcal{O}\big(M(M{-}1)(N_{\max}^2 + N_{\max}^3)\big),
\]
and is small in practice ($N_{\max}\!\le\!5$), making CCS a minor fraction of the overall pipeline.q


\section{Conclusion}
We presented CCS, a label-free, deployment-oriented monitoring  signal that converts test-time augmentation agreement into a  per-image proxy for detector behavior. By aggregating IoU overlaps across mild, non-geometric augmentations, CCS enables continuous monitoring and side-by-side comparison without ground-truth annotations. In controlled studies on Open Images and KITTI, CCS exhibits $>90\%$ directional congruence with F1-score, pPDQ, and OC-cost, alongside strong monotonic 
alignment and robustness across augmentation seeds and heterogeneous architectures. Qualitative consistency was further confirmed on COCO and BDD100K validation splits. Furthermore, we provided a simplified theoretical analysis that links the expected CCS to detection correctness under an idealized setting. Using a calibrated indifference margin, small discrepancies are treated as ties, emphasizing substantive disagreements and facilitating targeted inspection of unstable cases, supporting the use of CCS as a standalone monitoring signal in operational settings where labels are unavailable.


\bibliographystyle{IEEEtran}
\bibliography{references} 

\end{document}